\documentclass[12pt]{article}

\usepackage{arxiv}

\usepackage[utf8]{inputenc}
\usepackage[T1]{fontenc}
\usepackage{hyperref}
\usepackage{url}
\usepackage{booktabs}
\usepackage{amsfonts}
\usepackage{graphicx}
\usepackage{subfigure}
\usepackage{multirow}
\usepackage[numbers]{natbib}
\usepackage{enumitem}
\usepackage{array}

\newcolumntype{L}[1]{>{\raggedright\let\newline\\\arraybackslash\hspace{0pt}}m{#1}}
\newcolumntype{C}[1]{>{\centering\let\newline\\\arraybackslash\hspace{0pt}}m{#1}}
\newcolumntype{R}[1]{>{\raggedleft\let\newline\\\arraybackslash\hspace{0pt}}m{#1}}

\newcommand{\eqref}[1]{(\ref*{#1})}

\title{An evaluation framework for synthetic data generation models}

\date{}                    

\author{ 		
	\href{https://orcid.org/0000-0002-3996-3301}{\includegraphics[scale=0.06]{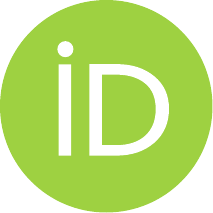}\hspace{1mm}Ioannis E.~Livieris}\thanks{Corresponding author}\\
	Novelcore,\\
	Athens, GR 185-34.\\
    \texttt{livieris@novelcore.eu} \\
    \And
	\href{https://orcid.org/0009-0000-0827-9729}{\includegraphics[scale=0.06]{orcid.pdf}\hspace{1mm}Nikos Alimpertis}\\
	Novelcore,\\
	Athens, GR 185-34.\\
	\texttt{alimpertis@novelcore.eu} \\
	\And
	\href{https://orcid.org/0000-0002-3449-2029}{\includegraphics[scale=0.06]{orcid.pdf}\hspace{1mm}George Domalis}\\
	Novelcore,\\
	Athens, GR 185-34.\\
	\texttt{domaliss@novelcore.eu} \\
	\And
	\href{https://orcid.org/0000-0001-7185-102X}{\includegraphics[scale=0.06]{orcid.pdf}\hspace{1mm}Dimitris Tsakalidis}\\
	Novelcore,\\
	Athens, GR 185-34.\\
	\texttt{tsakalidis@novelcore.eu}		
}

\renewcommand{\shorttitle}{An evaluation framework for synthetic data generation models}

\hypersetup{
    pdftitle={An evaluation framework for synthetic data generation models},
    pdfsubject={cs.CV, cs.LG, cs.AI},
    pdfauthor={Ioannis E. Livieris, Nikos Alimpertis, George Domalis and Dimitris Tsakalidis},
    pdfkeywords={Causal inference, XGBoost, treatment effect estimation, potential outcomes},
}

\begin{document}
    \maketitle

\begin{abstract}
Nowadays, the use of synthetic data has gained popularity as a cost-efficient strategy for enhancing data augmentation for improving machine learning models performance as well as addressing concerns related to sensitive data privacy.
Therefore, the necessity of ensuring quality of generated synthetic data, in terms of accurate representation of real data, consists of primary importance. In this work, we present a new framework for evaluating synthetic data generation models' ability for developing high-quality synthetic data. The proposed approach is able to provide strong statistical and theoretical information about the evaluation framework and the compared models' ranking. Two use case scenarios demonstrate the applicability of the proposed framework for evaluating the ability of synthetic data generation models to generated high quality data. The implementation code can be found in \url{https://github.com/novelcore/synthetic_data_evaluation_framework}. \\ \\
	*** This paper has been accepted for presentation at \textit{IFIP International Conference on Artificial Intelligence Applications and Innovations}. Cite: Livieris I.E., Alimpertis N., Domalis G., \& Tsakalidis D. (2024). An evaluation framework for synthetic data generation models. \textit{IFIP International Conference on Artificial Intelligence Applications and Innovations}.***
\end{abstract}

\keywords{Synthetic data generator\and evaluation framework\and tabular data\and statistical analysis.}

\section{Introduction}

In this data-driven era, the collection and acquisition of a sufficient amount of high-quality data play a vital role in enhancing decision-making and innovation.
Many real-world applications such as medicine and finance possess ethical and privacy restrictions, requiring that data confidentiality should be secured without sacrificing quality. As a result, the increasing need for privacy-preserving data as well as the lack of high-quality data constitute considerable drawbacks for the development of effective solutions in many real-world applications. The generation of artificially developed synthetic data offers a powerful tool for dealing with these drawbacks. Nowadays, the use of synthetic data has gained popularity as a cost-efficient strategy for data augmentation focusing on enhancing the performance of machine learning models while addressing sensitive data privacy concerns.


Mathematically, the task of synthetic data generation of a real dataset $\mathcal{D}_R$ requires the development of a synthesizer $S$ for developing the synthetic dataset $\mathcal{D}_S$. Dataset $D_R$ contains $N_c$ continuous features $\{c_1, c_2, \dots, c_{N_c}\}$ and $N_d$ discrete features $\{d_1, d_2, \dots, d_{N_d}\}$, where each feature is considered as a random variable, which follow an unknown, usually non-Gaussian, joint distribution. In the literature, a variety of approaches has been proposed ranging from traditional machine learning models such as Gaussian Mixture Models \cite{fruhwirth2019handbook} to the most elegant deep learning models such as Generative Adversarial Networks \cite{gui2021review}. These models offer researchers to choose between probabilistic modeling and adversarial training dynamics.

The process of evaluating synthetic data is generally considered as a complex task and it is usually based on the employment of evaluations tests, which can be divided in two main categories: \textit{univariate} and \textit{multivariate}. Univariate tests provide valuable insights into individual variables as well as their pairwise relationships. However, they fail capture the intricate multivariate relationships between the features. In contrast, multivariate tests are able to provide a deeper understanding of the underlying patterns and structures, by studying the dependencies and interactions among variables and assess the joint distribution of three or more variables.

However, in most cases, the evaluation tests cannot be theoretically compared \cite{espinosa2023quality} while frequently they have been found to generate conflicting conclusions \cite{dankar2022multi}, which implies that the direct comparison of synthetic data generators is considered a considerable complex and difficult task. The traditional approach for addressing this difficulty is to rank the models by counting the number of tests, which each model exhibited the top score \cite{boehmer2022quantitative} or by calculating the average of the normalized scores, to obtain a quality score for each model \cite{chundawat2022tabsyndex}. Hernadez et al. \cite{hernandez2022synthetic} conducted a systematic review on the evaluation of synthetic data generation models and concluded that there is no universal method or metric to effectively compare the performance of evaluated approaches. In addition, in case the real data is unlabeled, the complexity of the evaluation process is increased, since the generated synthetic data cannot be used as augmentation data for addressing classification and/or regression tasks.

In this work, we propose a new framework for evaluating and compare the performance of synthetic data generation models relative to the quality of the generated synthetic datasets. The proposed framework is based on the employment of several multivariate evaluation tests for measuring the quality of generated synthetic data. Based on the tests results, the models are ranked using Friedman Aligned-Ranks test \cite{hodges2012rank} and the existence of significant differences in their performance is examined using Finner post-hoc test \cite{finner1993monotonicity}. Therefore, the proposed approach is able to provide strong statistical and theoretical evidence about the models' ranking and the overall evaluation framework. Additional advantages of the proposed framework are its flexibility and adaptivity in sense that new tests can be easily included and it can be employed for evaluating the quality performance of any generated synthetic dataset. It is worth mentioning that in this work, we examine the case where the real data is unlabeled, which

\noindent considerably increases the complexity of the evaluation process. The experimental analysis was based on the application of proposed framework for measuring the data quality of state-of-the-art synthetic data generators on two tabular real-world datasets. In summary, the main contributions of this research are:
\begin{itemize}
	\item We propose a new framework for ranking synthetic data generation models relative to their ability to generate high-quality synthetic data.
	
	\item The proposed approach is able to provide strong statistical and theoretical evidence about the evaluation framework and models' ranking.
	
	\item The introduced framework has the advantages of flexibility and adaptivity in sense that new evaluation tests can be easily integrated as well as it possesses minimum assumptions requirements for its application.
	
	\item The use case scenarios highlight the applicability of the proposed framework as well as its ability to evaluate the quality of generated synthetic data and obtain safe conclusions.
\end{itemize}

The remainder of this paper is organized as follows: Section \ref{Sec:2} presents a brief survey of recent studies for the evaluation of synthetic data generation. Section \ref{Sec:3} presents the state-of-the-art models for synthetic generation of tabular data. Section \ref{Sec:4} presents the proposed evaluation framework and its main components. Finally, Section \ref{Sec:5} presents the use case scenarios, while Section \ref{Sec:6} summarizes the conclusions of this work and proposed some directions for future research.

\section{Related work}\label{Sec:2}

The generation of high-quality synthetic data has become integral for addressing challenges in many real-world applications \cite{nikolenko2021synthetic}. Nowadays, synthetic data plays a major role in augmenting training datasets and facilitating responsible data sharing. The effectiveness of synthetic data generation methods is focused on their ability to model the complex multivariate distributions, replicate the features' statistical properties as well as inherent the complexities presented in the real datasets; hence, the necessity for a robust evaluation methodology has become essential. In the sequel, we provide a brief description of some rewarding research studies for synthetic data generators evaluation.

Bourou et al. \cite{bourou2021review} performed a review study of tabular data synthesis using GANs for intrusion detection systems. Specifically, the authors studied the performance of three of the most popular GAN-based model, namely CTGAN, CopulaGAN, and TableGAN to generate synthetic tabular datasets on the NSL-KDD dataset. The evaluation process included a visual procedure based on the correlation matrix between the features, several statistical metrics including Kolmogorov–Smirnov test and Chi-squared test as well as Machine learning efficacy test. Based on the experimental analysis, the authors stated that CopulaGAN and CTGAN presented the best overall performance.

Hernandez et al. \cite{hernandez2022synthetic} presented a systematic review for analyzing approaches in synthetic tabular data generation, focusing on the healthcare applications. In their study, special attention was payed to GANs models, which were employed for two use case scenarios: data augmentation and privacy-preservation. In addition, the authors conducted a six-step procedure for reviewing several methodologies based on univariate and multivariate statistical tests for comparing synthetic data generation models. Based on their comprehensive analysis, they concluded that no universal metric or methodology exists for the models' evaluation relative to the quality of the generated data.

Figueira et al. \cite{figueira2022survey} conducted a comprehensive 
survey for synthetic data generation models. The authors highlighted the advantages of each approach as well as its limitations, providing special attention on the class of GANs models. Finally, they presented the most widely used techniques  for evaluating the quality of synthetic data, which were divided in visual-based such as correlation matrix and graphical representation, and in metrics-based such GAN-train, GAN-test, descriptive statistics, $\alpha$-precision and $\beta$-recall.

Dankar et al. \cite{dankar2022multi} presented a comprehensive evaluation of several tabular synthetic data generators. The evaluation process focused on comparing the models' ability to preserve: population, attribute, bivariate and application fidelity.  For this purpose, the authors selected a representative test from each category, which was employed for the evaluation of four data synthesizers on 19 datasets with different characteristics (number of instances, feature counts, feature types). Finally, the linear correlations of all selected test were examined for identifying possible pairwise relationships. Based on their analysis, the authors states that the use of several evaluation tests provide conflicting conclusions, making the direct comparison of synthetic data generation models a rather challenging task.

In this work, we propose a new framework for evaluating and ranking the synthetic data generation models' ability to generate high-quality artificial data. In contrast to previous approaches, it is able to provide strong theoretical and statistical information about models' ranking relative to the quality of the  generated synthetic data. Its advantages are its flexibility and adaptivity in sense that new evaluation tests and metrics can be easily integrated as well as it possesses minimum assumptions requirements for its application.\vspace{-.2cm}

\section{Synthetic data generation methods for tabular data}\label{Sec:3}

In this section, we briefly present some of the effective and widely used models for generating tabular data. It is worth mentioning that traditional approaches such as Synthetic Minority Over-sampling Technique \cite{chawla2002smote} and Adaptive Synthetic Sampling \cite{he2008adasyn} were not included, since they require labeled data for their application. Nowadays, Gaussian Mixture Models (GMM) and Generative Adversarial Networks (GANs) constitute the most widely used approaches proposed in the literature for generating synthetic data. These classes of models offer researchers to choose between probabilistic modeling based on the specific requirements of tabular data at hand and adversarial training dynamics. Another interesting approach for generating synthetic data is based on \textit{Gaussian Copula} \cite{li2020sync}, which is based on statistical methods for modeling features dependencies. It is able to capture linear dependencies in multivariate data by modeling the features' joint distribution given their marginal distributions. In contrast to GMM and GANs, which are dedicated to create synthetic data, Gaussian Copula is not a generative model in sense of creating new instances directly, but is often used in synthetic data generation by combining it with marginal distributions to generate synthetic instances with similar dependency structures.

\textit{Gaussian Mixture Models} \cite{fruhwirth2019handbook} are characterized by their ability to model complex multivariate distributions and offer an effective approach for synthetic data generation. GMM are based on a probabilistic framework for modeling complex data distributions by combining multiple Gaussian components. In detail, these models learns the distribution of the input data, focusing on explicitly capturing the interdependence between variables using mixture components for generating synthetic samples, which closely resemble the original dataset. The versatility of GMM lies in their ability to capture diverse patterns and dependencies within tabular data, making them well-suited for many real-words applications.

\textit{Generative Adversarial Networks} \cite{gui2021review} have emerged as a state-of-the-art models for synthetic data generation, demonstrating significant capabilities in preserving the statistical properties of original datasets by maintaining both the marginal distributions of individual features and complex inter-feature dependencies. \textit{Conditional Tabular Generative Adversarial Network} (CTGAN) \cite{xu2019modeling} constitutes a GAN-based model, which is dedicated for synthesizing tabular data. It is based on a sophisticated architecture to address the challenges of modeling tabular data. Specifically, for dealing with multimodal and non-Gaussian distribution, CTGAN employs a variational Gaussian mixture model (mode-specific normalization) for converting continuous values into a bounded vector, which constitutes a representation suitable for neural networks. An additional advantage of CTGAN is that is able to focus on specific features for generating synthetic data with desired attributes. This capability proves invaluable in scenarios where maintaining the integrity of specific data distributions and relationships is crucial. 

\textit{Table Variational Auto-Encoder} (TVAE) \cite{xu2019modeling} is a synthetic data generator model, which is composed by an encoder and a decoder. It maps tabular data to a low-dimensional latent space representation, which follow a arbitrary probability distribution, while the decoder leverages the latent space for generating and re-constructing synthetic data. This model was originally proposed for evaluating the synthetic data generation capabilities of the GAN-based models by using a non-GAN generator model.

\textit{Copula Generative Adversal} (CopulaGAN) model \cite{espinosa2023quality} constitutes an efficient variation of CTGAN, which exploits the cumulative distribution function-based transformation, applied through Gaussian Copula \cite{kamthe2021copula}. Copulas are based on probability theory aiming at describing the inter-correlation between the variables. During the training process, a reversible data transformation is employed for handling null and non-numerical data, and create a fully numerical representation, which is used by the model is able to learn the probability distributions of each feature. The advantage of this approach is that CopulaGAN is able to learn the data distribution and correlation between the features accurately \cite{bourou2021review}.

Summarizing, GANs have been established in the literature by their ability of generating high-quality synthetic data. Nevertheless, these models are computationally expensive and suffer from ``\textit{model collapse}'' problem, which implies that they repeatedly generate a limited set of outputs; therefore, failing to produce diverse and meaningful samples. In other words, a GAN-based model collapses to specific regions of the input space, producing nearly identical samples regardless of input variations and random noise. In contrast, although GMM have the advantages of scalability and considerably less computational effort requirements compared to GANs, the quality of the generated data is sometimes inferior to those of GANs since they possess limited ability to capture nonlinear relationships in case some features do not follow a Gaussian distribution. Hence, we are able to conclude that no universal model exists and its selection is based on the use case scenario.\vspace{-.2cm}

\section{Proposed synthetic dataset evaluation framework}\label{Sec:4}

Algorithm 1 presents the proposed evaluation framework, which takes as input a real dataset $D_R$ and a set of synthetic generation models $\mathcal{S} = \{S_1, S_2, \dots, S_N\}$; while the output is the selected synthetic dataset $D_{S_{opt}}$. 

Initially, the model $S\in\mathcal{S}$ is employed for generating the synthetic dataset $D_{S}$, aiming at preserving the patterns, structural characteristics and relationships of the real dataset $D_R$ (Steps 1-2). Then, in Step 3, the \textit{Diagnostic Validity test} is applied, for examining the basic data validity and data structure issues of the synthetic dataset $D_{S}$ (i.e. unique and non-null primary keys, continuous values adhere to the min/max range in the real data, and  discrete values must adhere to the same categories as the real data). In case where any of these conducted examinations is violated then the dataset $D_{S}$ is not further considered for evaluation. Next, in Steps 5-8, \textit{Wasserstein-Cramer's V}, \textit{Novelty}, \textit{Domain classifier} and \textit{Anomaly detection} evaluation tests are applied for measuring the quality of each generated synthetic dataset, and a corresponding score is calculated for each test. 
These tests probably constitute the most widely used evaluation tests in the literature \cite{hernandez2022synthetic,bourou2021review,figueira2022survey} for comparing synthetic data generator models.\vspace{-.1cm}

\begin{itemize}
	\item \textit{Wasserstein-Cramer's V test} \cite{figueira2022survey} constitutes one of the simplest and widely used synthetic data evaluation tests. It is performed on each feature separately by calculating the Wasserstein distance, in case the feature is numerical or Cramer's V in case the feature is categorical. In general, a relative small score indicates that the quality of the synthetic dataset for that particular feature is high. Although this test has the advantages of requiring minimum computational cost it is limited by the fact that it is univariate. For addressing this issue, researchers proposed for each model to calculate the mean for of all calculated features \cite{boehmer2022quantitative,chundawat2022tabsyndex}. However, the mean value may lead to misleading conclusions, since a small number of values may tend to dominate the results \cite{kiriakidou2024mutual,livieris2022dropout}. Hence, in this work, we consider a different approach and employ Friedman Aligned Ranking test \cite{hodges2012rank} for ranking the models and obtaining the corresponding scores. This approach provides a statistical approach for calculating a representative score about quality of the datasets based on the calculated scores for this test.
	
	\item \textit{Novelty test} \cite{herurkar2023cross} aims to assess the synthetic data models' diversification capability by measuring their ability to introduce novel instances not present in the original real data. The score for each model is calculated as the ratio of unmatched instances contained in each generated synthetic dataset. Notice that two instances $s$ and $r$ from synthetic and real datasets, respectively, are considered a match, in case all the individual values $s_i$ in the synthetic instance are relative close to the real instances $r_i$, i.e. $|s_i - r_i|<\alpha$, where $\alpha$ is a pre-defined threshold.
	
	\item \textit{Domain classifier test} \cite{spadotto2021unsupervised} focuses on training a binary classification model to discriminate between instances from real and synthetic datasets. If the model is able to confidently identify which instances belong to the real and which ones to the synthetic dataset, it implies that the two datasets are significantly different; hence, the synthetic dataset is considered of poor quality. For this test, the score for each model is defined as the mean Area Under the ROC Curve (AUC) \cite{canbek2017binary} of the domain classifier obtained from each iteration of the stratified cross validation process. Notice that in case AUC tends to 0.5 implies that the synthetic instances cannot be distinguished from the instances of the real dataset; while in contrast, in case AUC tends to 1 implies that the synthetic dataset is of poor quality.
	
	\item \textit{Anomaly detection test} \cite{llugiqi2022empirical} is based on studying the patterns and structure of the real dataset and examine the synthetic dataset for detecting anomalies. The motivation behind this test involves assessing how well the synthetic dataset conforms to the expected patterns of the real dataset by identify any anomalies or deviations from the original data distribution. For calculating the score for each model, initially, the isolation forest \cite{lesouple2021generalized} is trained on the real dataset and then it is applied on the synthetic dataset for assigning a score to each synthetic data point, indicating the degree of abnormality. Then, each model's score is defined as the mean score assigned to each instance contained in the synthetic dataset. Notice that an increase in the percentage of anomalies indicates a deviation from the expected normal behavior, which implies the poor quality of the synthetic dataset.
\end{itemize}

After, the evaluation test scores for each model are calculated, a statistical analysis is applied, using Friedman Aligned-Ranks (FAR) test \cite{hodges2012rank} and post-hoc Finner test \cite{finner1993monotonicity}. The former is used for ranking the synthetic data generation models while the latter is used for examining the existence of significant differences in the quality of the generated data \cite{kiriakidou2024mutual}, respectively (Step 9). 
The statistical analysis examines the rejection or the acceptance of the hypothesis $$
H_0:\{\textnormal{the quality of all generated synthetic datasets is similar}\},
$$
and provides statistical evidence about the presence of significant variations in their calculated scores, without assuming a specific distribution of the performance scores \cite{garcia2010advanced,livieris2019detecting}.
Under the null hypothesis $H_0$, the FAR test statistic $F_{AR}$ is defined by\vspace{-.2cm}
$$
F_{AR} = \frac{(k-1)\left[\sum_{i=1}^N\hat{R}_i^2-(kn^2/4)(kn+1)^2\right]}{([kn(kn+1)(2kn+1)]/6)-\displaystyle\frac{1}{k}\sum_{j=1}^n\hat{R}_j^2},\vspace{-.2cm}
$$
where $\hat{R}_i$ is equal to the rank total of the $i$-th model and $\hat{R}_j$ is equal to the rank total of the $j-$th test. The test statistic $F_{AR}$ is compared with $\chi^2$ distribution with ($k-1$) degrees of freedom and $\alpha=5\%$ significance in order to examine whether $H_0$ can be rejected. It is worth mentioning that two advantages of FAR test is that (i) it uses rankings rather than looking directly at the scores themselves and (ii) it does not require the commensurability of the measures across different tests as well as does not assume the normality of the sample means; hence, it is robust to outliers. Notice that in case $H_0$ failed to be rejected the employment of post-hoc Finner test is necessary in order to examine the existence of significant differences among the synthetic data quality. Finner post-hoc test adjusts the value of the level of significance $\alpha$ in a step-down manner and rejects $H_0$ to $H_i$ if $i$ is the smallest integer so that $p_i = 1-(1-\alpha)^{(k-1)/i}$. Morevoer, the adjusted Finner $p$-value is defined by
$$
APV_i = \min\{v;1\},
$$
where $v=max\{1-(1-p_j)^{(k-1)/j}:\, 1\le j\le i\}$, $p_i$ are the ordered $p$-values, with $p_1\le p_2\le \dots, p_k$ and $H_1,H_2,\dots,H_k$ be the corresponding alternative hypotheses. Finally, the generated dataset of the model, which reported the best overall performance based on the conducted statistical analysis, constitutes the output of the proposed framework (Step 10).

\noindent\rule{13cm}{0.4pt}\\
\noindent{}\textbf{Algorithm 1:}\vspace{-.2cm}
\begin{description}
	\item[Inputs:]
	\item[] \quad $D_R$: Real dataset
	\item[] \quad $\mathcal{S}$: Set of synthetic data generation models.
	\item[Output:]
	\item \quad $D_{S_{opt}}$: Selected synthetic dataset.
	\clearpage
	\setlength\itemsep{.3em}
	\item[Step 1:] for $S$ in $\mathcal{S}$ do
	\item[Step 2:] \quad Create dataset $D_S$ using synthesizer $S$ on $D_R$.
	\item[Step 3:] \quad Examine the validity of dataset $D_S$ using the \textit{Diagnostic Validity}
	\item[]\hspace{1.5cm} \textit{test}. In case, any violations are identified, $D_S$ is no further considered 
	\item[]\hspace{1.5cm} for evaluation.
	\item[Step 4:] end for
	\item[Step 5:] Apply \textit{Wasserstein/Cramer's V test} and calculate the 
	\item[]\hspace{1.3cm}corresponding scores for each evaluated model.
	\item[Step 6:] Apply \textit{Novelty test} and calculate the corresponding scores for each 
	\item[]\hspace{1.3cm}evaluated model.
	\item[Step 7:] Apply \textit{Domain Classifier test} and calculate the corresponding scores
	\item[]\hspace{1.3cm}for each evaluated model.
	\item[Step 8:] Apply \textit{Anomaly Detection test} and calculate the corresponding scores
	\item[]\hspace{1.3cm}for each evaluated model.    
	\item[Step 9:] Apply a statistical analysis (FAR \& Finner tests) based on calculated
	\item[]\hspace{1.3cm}evaluation scores of each model.
	\item[Step 10:] Return $D_{S_{opt}}$
\end{description}\vspace{-.4cm}
\noindent\rule{13cm}{0.4pt}\vspace{.1cm}


\section{Experimental Analysis}\label{Sec:5}

In this section, we apply the proposed framework for evaluating the performance of the synthetic data generation models: Gaussian Copula, GMM, CTGAN, TVAE and CopulaGAN and measuring the quality of the generated synthetic data. For evaluation purposes, we employed two datasets from different real-world application domains:

\setcounter{footnote}{0}

\begin{itemize}
	\item \textbf{Travel Review Ratings}\footnote{\url{https://doi.
			org/10.24432/C5C31Q}}. This dataset is based on users' average rating information, captured from Google reviews, on 24 types of attractions across Europe. Google user rating spans from Terrible (1), Poor (2), Average (3), Very Good (4) and Excellent (5) with the average user rating computed for each category.
	
	\item \textbf{Obesity risk dataset}\footnote{\url{https://doi.
			org/10.34740/kaggle/dsv/7009925}}. This dataset concerns the estimation of obesity levels with diverse eating habits and physical condition. The data was collected from anonymous users (Mexicans, Peruvians and Colombians) using an online survey, which replied to a variety of questions concerning physical conditions, eating habits and personal/demographic information.
\end{itemize}

Similar to previous researchers \cite{espinosa2023quality,dankar2022multi,hernandez2022synthetic}, \textit{Wasserstein-Cramer's}, \textit{Novelty}, \textit{Domain classifier} and \textit{Anomaly detection} evaluation tests were applied for measuring the quality of generated datasets for each use case and examine if their interpretation is able to reveal the best synthesizer model. Our aim is the highlight the difficulty of identifying the best model in terms of quality of the generated data as well as the necessity of employing the proposed framework for obtaining concrete results. In addition, we provide 2D visualizations of the real dataset along with each generated synthetic dataset for enhancing the interpretability of the evaluation framework but also for offering a direct and accessible comparison between the generated synthetic datasets and the original real dataset. 

In our experiments we used only 200 instances from each dataset for fitting all the models, while the evaluation was conducted on based on 1000 generated synthetic instances from each model.\footnote{The models parameters for each use-case as well as the implementation code can be found in \url{https://github.com/novelcore/synthetic_data_evaluation_framework}} It is worth mentioning that the number of the generated instances did not affected the conclusions made from our analysis.

\subsection{Use case: Travel Review Ratings}

Table~\ref{Table:Travel results} presents the scores of the evaluation tests for all synthetic data generation models. Notice that the top score for each test is highlighted with bold. Clearly, no model presents clear superiority since the tests  generate conflicting conclusions with GMM and Gaussian Copula exhibit the best score in two cases. Therefore, the use of the proposed framework is considered necessary for evaluating the models' performance.

Table \ref{Table:Travel} presents the statistical analysis from the application of the proposed evaluation framework. Friedman statistic $F_{AR}$ with 4 degrees of freedom is equal to 5.675, while the $p$-value is equal to 0.22, which suggests that the post-hoc test should be applied in order to examine the existence of significant differences among the models' performance. GMM exhibits the highest probability-based ranking, followed by TVAE and Gaussian Copula. Moreover, Finner test suggests that there are significant differences between GMM and the rest models in terms of quality of the generated data.

\begin{table}[!ht]
	\setlength{\tabcolsep}{5pt}
	\renewcommand{\arraystretch}{1.}
	\centering %
	\begin{tabular}{l|ccccc}
		\toprule
		Evaluation test        & GaussianCopula &    GMM     &  CTGAN   &   TVAE   & CopulaGAN \\ 
		\midrule
		Wasserstein-Cramer's V &     14.55      & \bf{11.0}  &   34.2   &   26.0   &   41.75   \\
		Novelty                &    \bf{0.0}    &   0.001    &  0.001 & \bf{0.0}   & \bf{0.0}  \\
		Domain classifier      &     0.783      & \bf{0.730} &  0.767   &  0.734   &   0.799   \\ 
		Anomaly  detection     &    \bf{0.0}    &   0.001    & \bf{0.0} &  0.001   &   0.013   \\
		\bottomrule
	\end{tabular}
	\caption{Score on evaluation tests of Gaussian Copula, GMM, CTGAN, TVAE and CopulaGAN models for travel review ratings dataset}\label{Table:Travel results}
%
	%
	\setlength{\tabcolsep}{10pt}
	\renewcommand{\arraystretch}{1.}
	\centering %
	\begin{tabular}{l|ccc}
		\toprule
		Model    & Friedman & \multicolumn{2}{c}{\underline{Finner post-hoc test}}\\
		& Ranking  & $APV_i$ & $H_0$\\
		\midrule
		GMM              & 6.125   &     $-$    &     $-$      \\
		Gaussian Copula  & 8.5     &  0.017548	&    Rejected  \\
		TVAE             & 9.5     &  0.000098  &    Rejected  \\
		CTGAN            & 12.75   &  0.000000  &    Rejected  \\
		CopulaGAN        & 14.63   &  0.000000	&    Rejected  \\
		\bottomrule 
	\end{tabular}
	\caption{Statistical analysis provided by the proposed evaluation framework for travel review ratings dataset}\label{Table:Travel}
\end{table}\vspace{-.5cm}

\begin{figure}[!ht]
	\centering
	\subfigure[CTGAN]{\includegraphics[width=.49\textwidth]{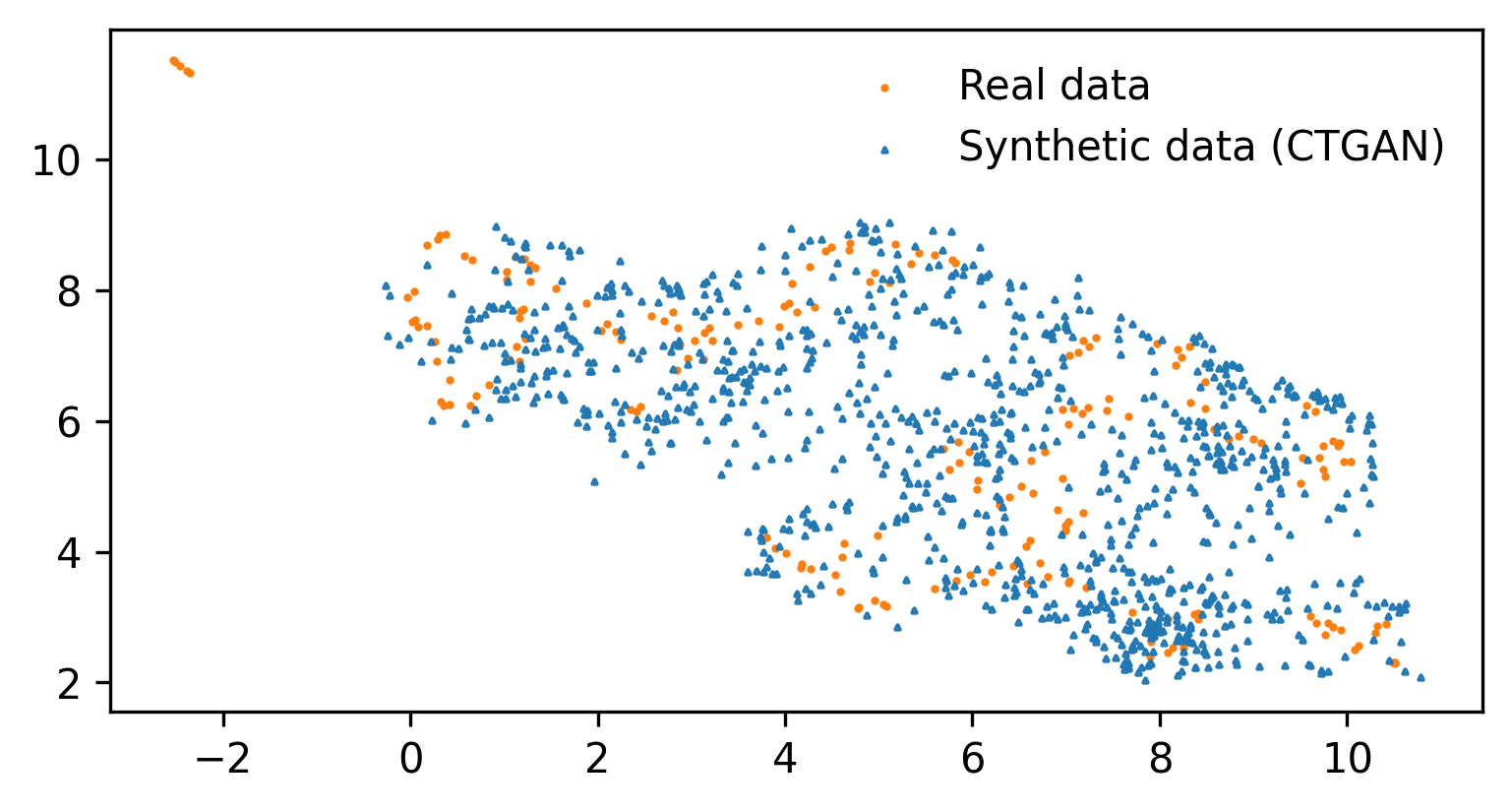}}
	\subfigure[TVAE]{\includegraphics[width=.49\textwidth]{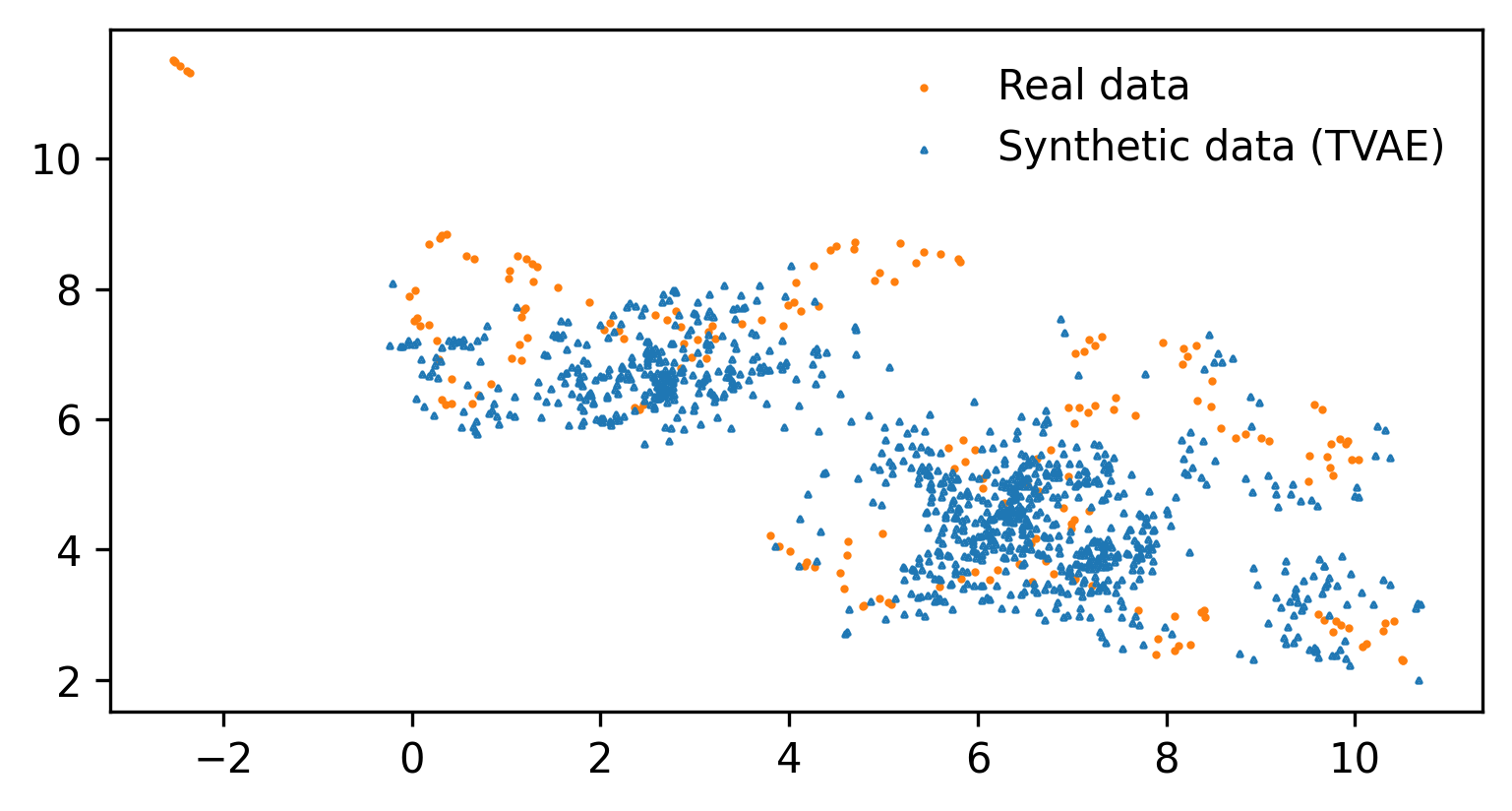}}
	\subfigure[CopulaGAN]{\includegraphics[width=.49\textwidth]{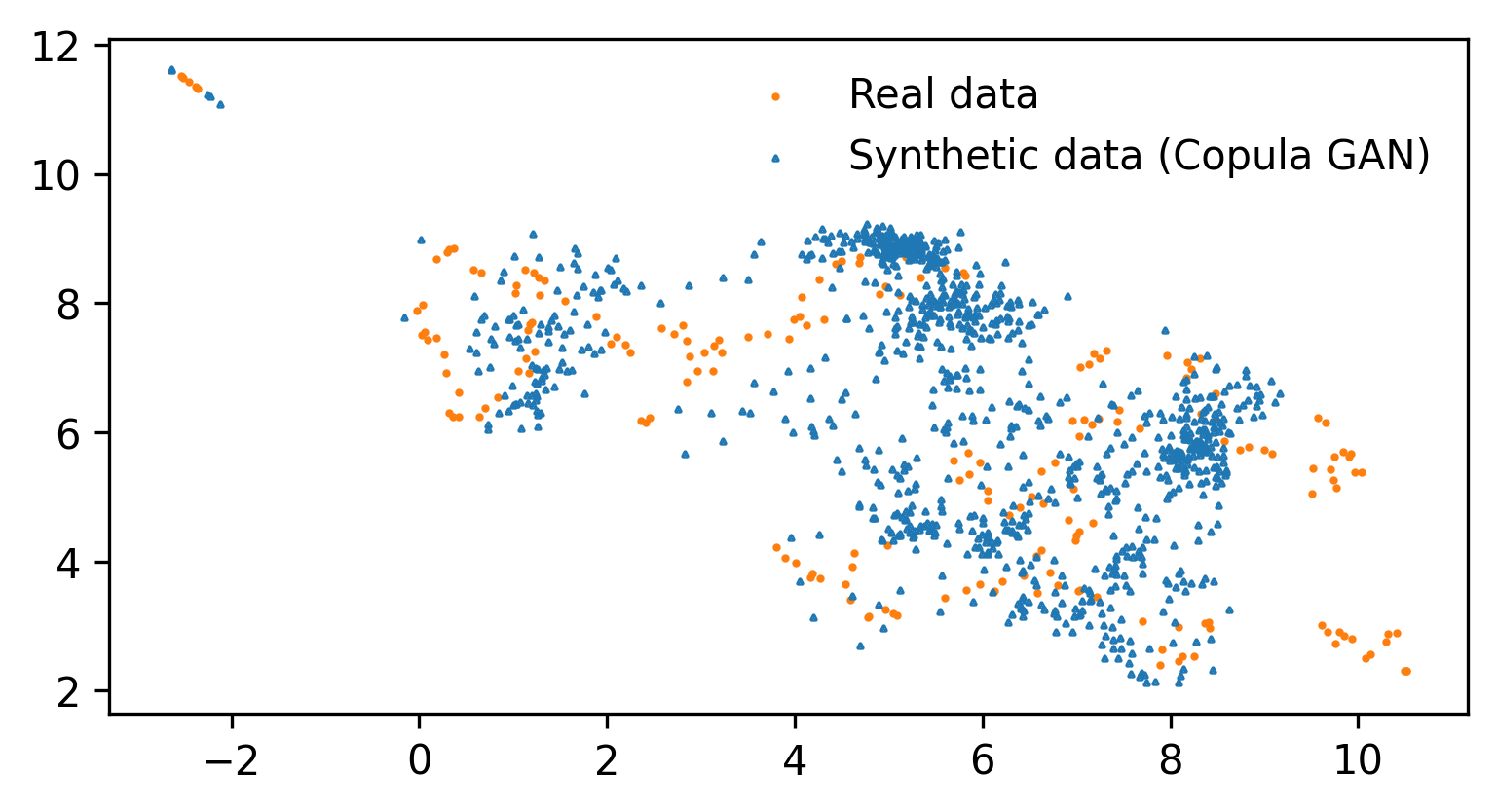}}
	\caption{2D visualization of generated synthetic data for travel review ratings dataset}\label{Fig:Travel}
\end{figure}

Figure \ref{Fig:Travel} presents a 2D visualization of real dataset along with each generated synthetic dataset. GMM produces synthetic data that are more representative of the entire dataset since they are more uniformaly distributed in sense that the generated data points are generated in a way that mimics the global characteristics of the real data. In contrast, Gaussian Copula, CTGAN, TVAE and CopulaGAN may introduce bias and inaccuracies since they tend to overly focus on certain regions of the data space, which could lead to obtaining a less reliable synthetic dataset.

\begin{figure}[!ht]
	\centering
	\subfigure[Gaussian Copula]{\includegraphics[width=.49\textwidth]{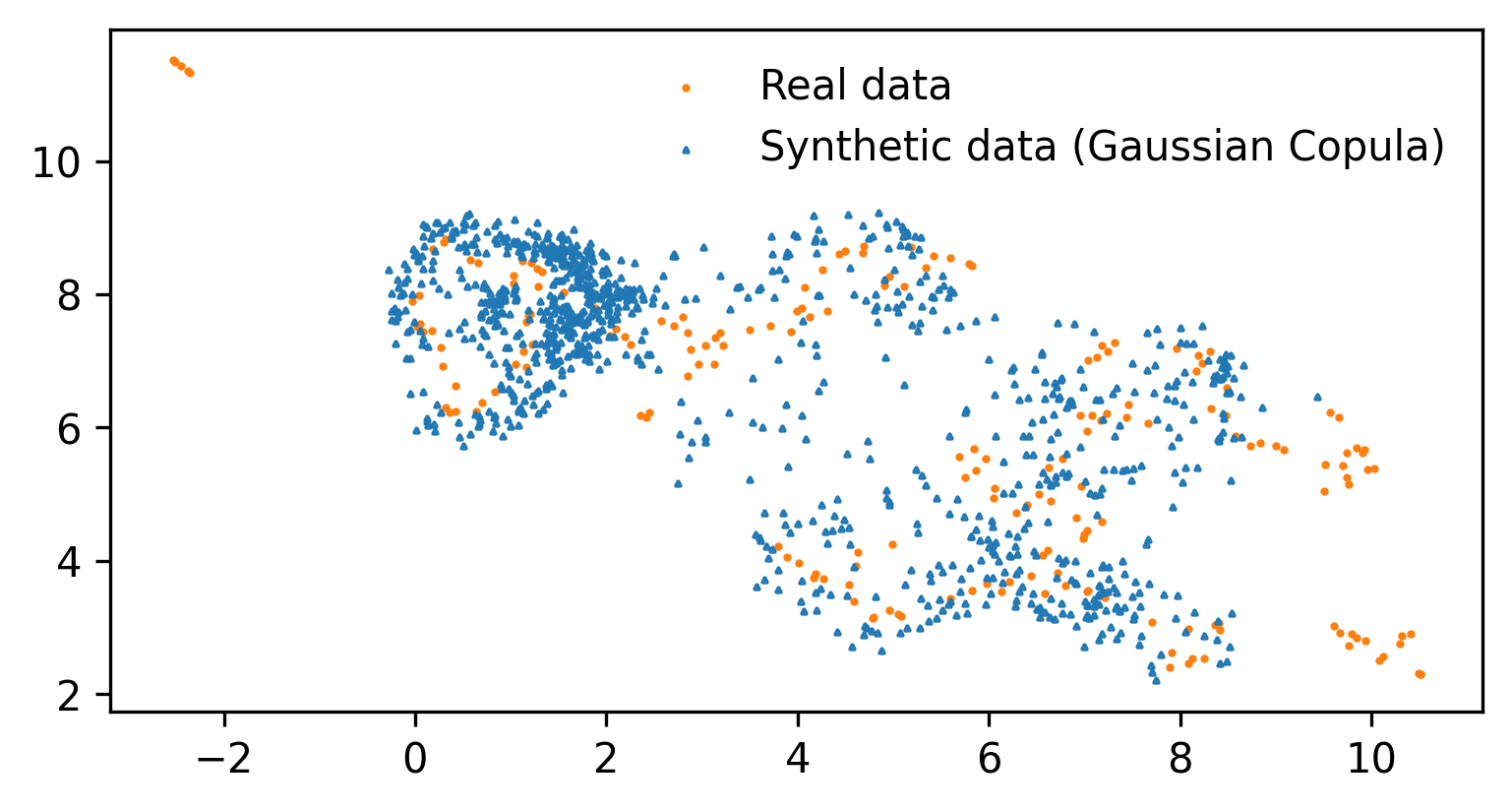}}
	\subfigure[GMM]{\includegraphics[width=.49\textwidth]{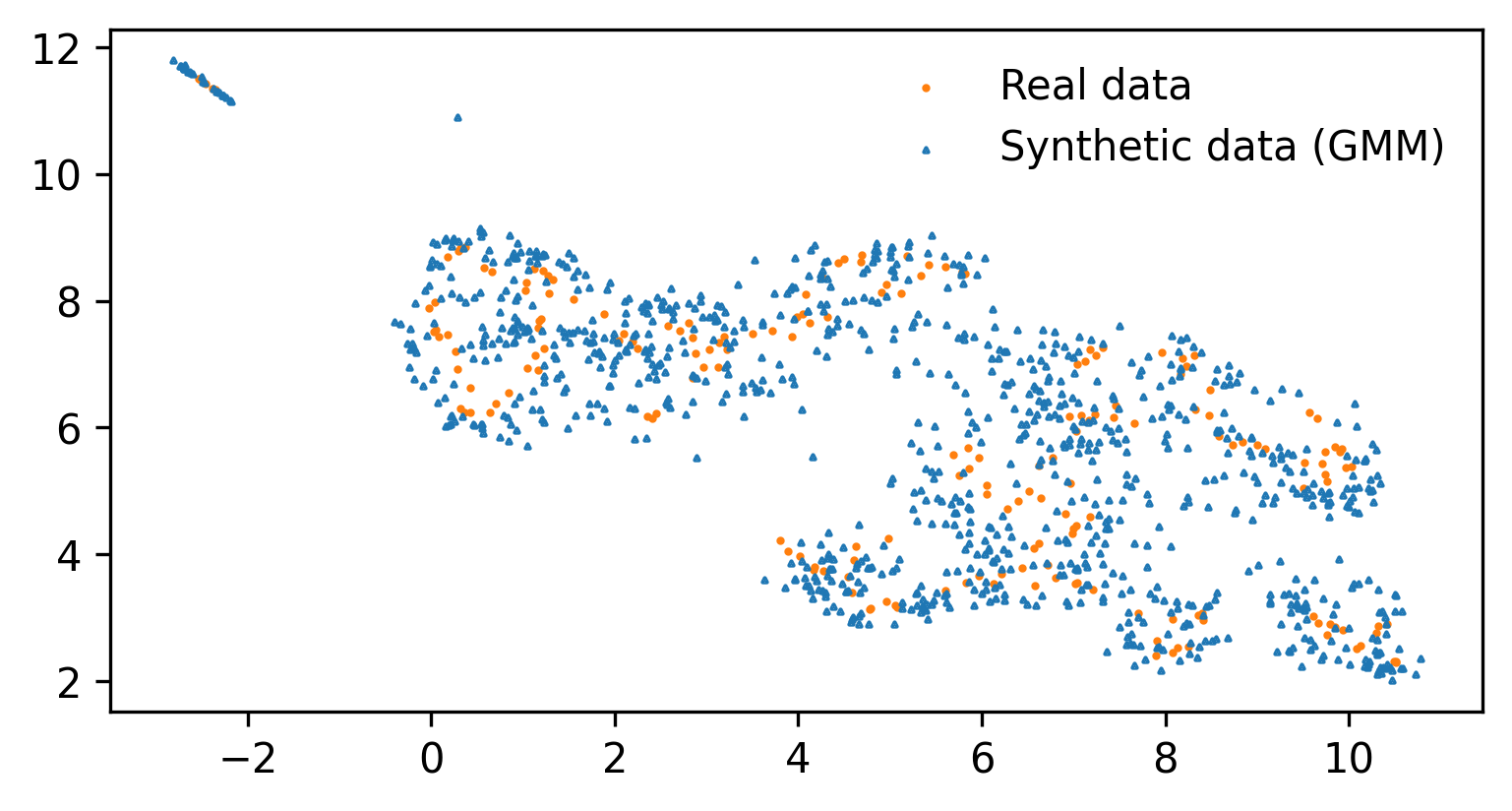}}
\end{figure}

\clearpage

\subsection{Use case: Obesity risk dataset}

Table~\ref{Table:Obesity results} presents the scores of the evaluation tests for Gaussian Copula, CTGAN, TVAE and CopulaGAN models while the best score for each test is highlighted with bold. CTGAN seems to present the best overall performance followed by TVAE since it exhibits the best scores in three cases. 

Table \ref{Table:Obesity} presents the statistical analysis of Gaussian Copula, TVAE, CTGAN, anc CopulaGAN. Notice that GMM was not included since it cannot be applied on mixed-type datasets. Friedman statistic $F_{AR}$ with 3 degrees of freedom is equal to 3.339, while the $p$-value is equal to 0.34. This suggests that Finner post-hoc test should be applied in order to examine the existence of statistical significant differences relative to the evaluated models' ability to generate quality data. CTGAN presents the best performance, reporting the highest probability-based ranking, followed by TVAE. In addition, Finner test suggests that there are significant statistical differences between the quality of the generated data by CTGAN and the corresponding data by the Gaussian Copula and CopulaGAN but not with those generated by TVAE.

\begin{figure}[!ht]
	\subfigure[Gaussian Copula]{\includegraphics[width=.49\textwidth]{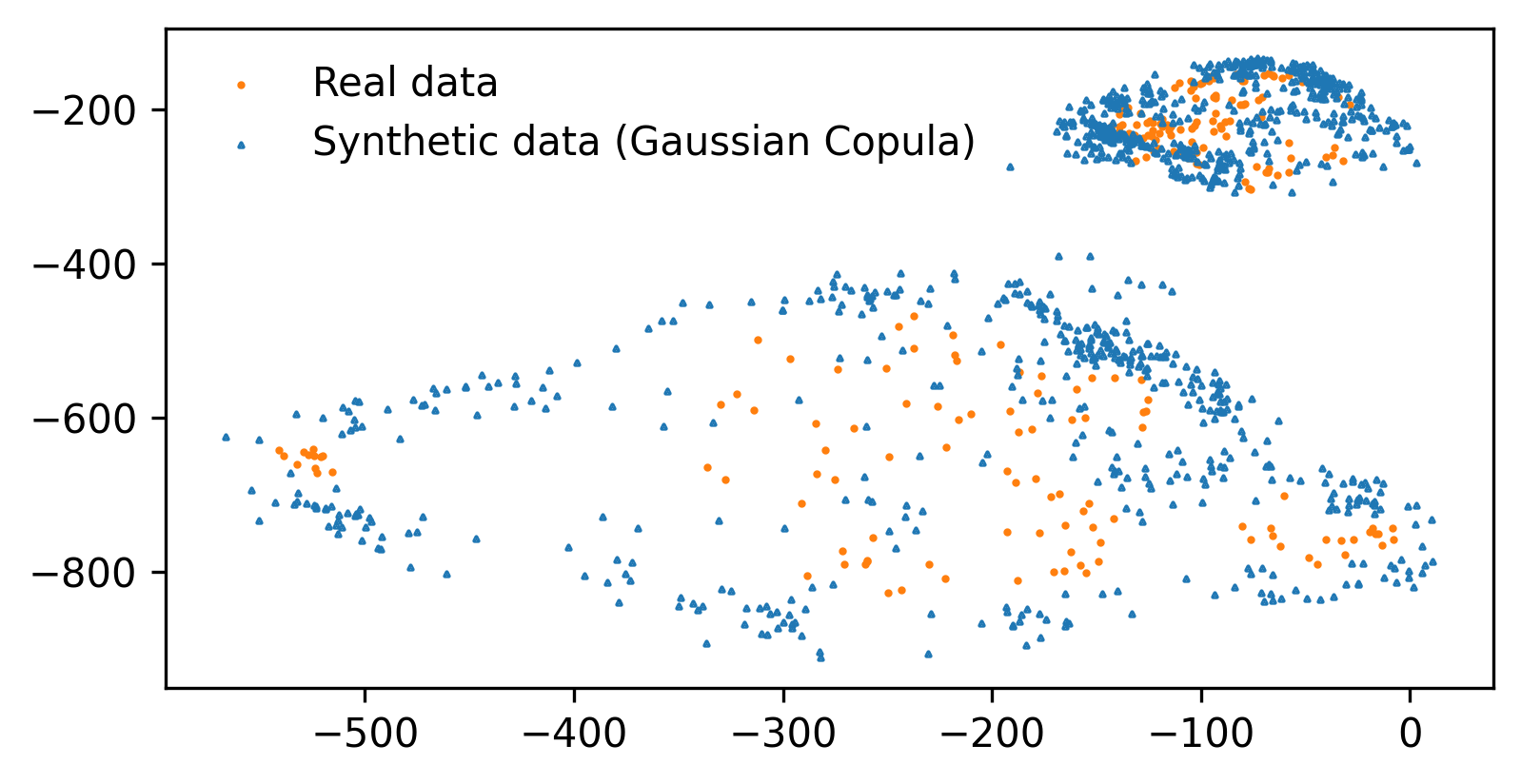}}
	\subfigure[CTGAN]{\includegraphics[width=.49\textwidth]{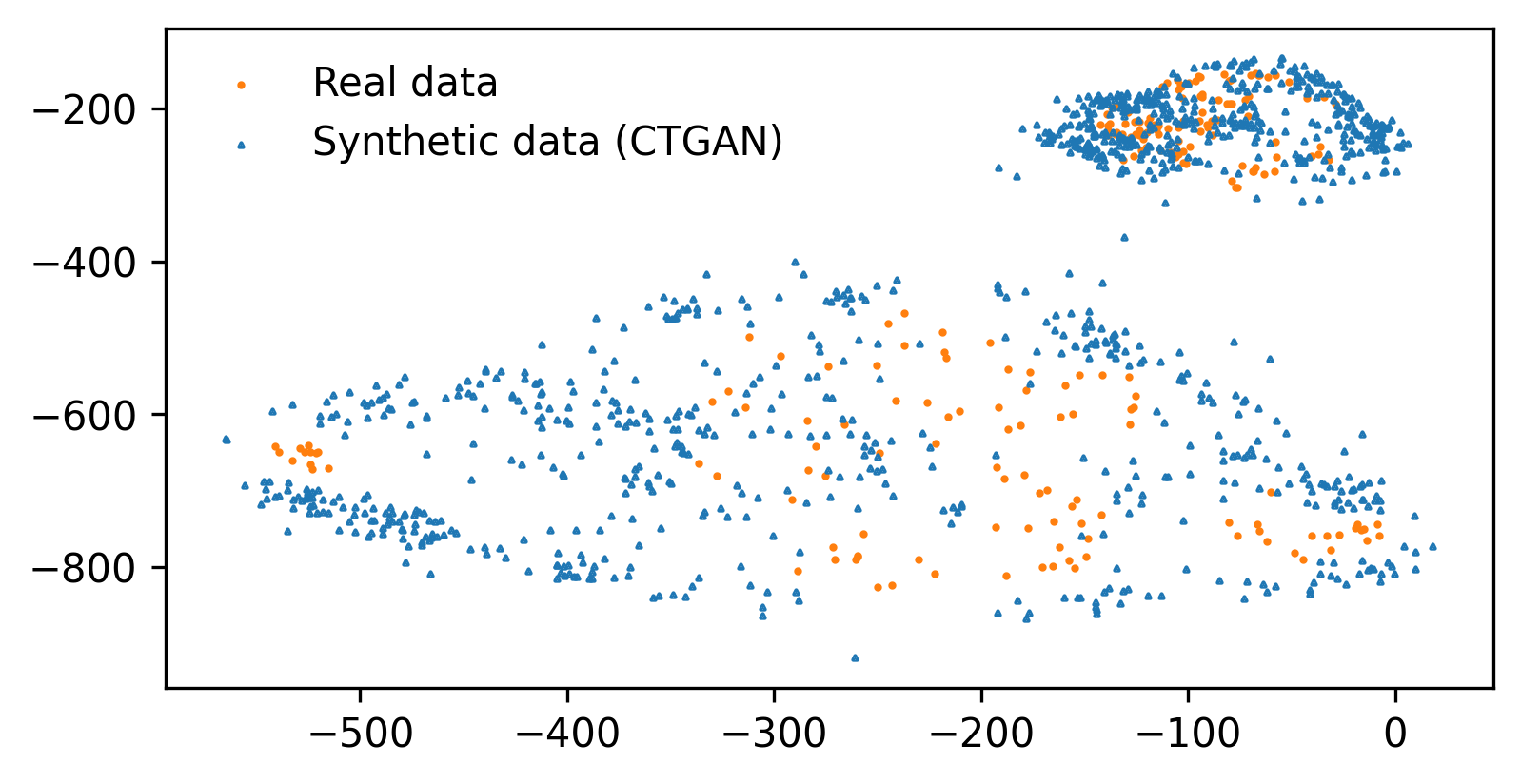}}
	\subfigure[TVAE]{\includegraphics[width=.49\textwidth]{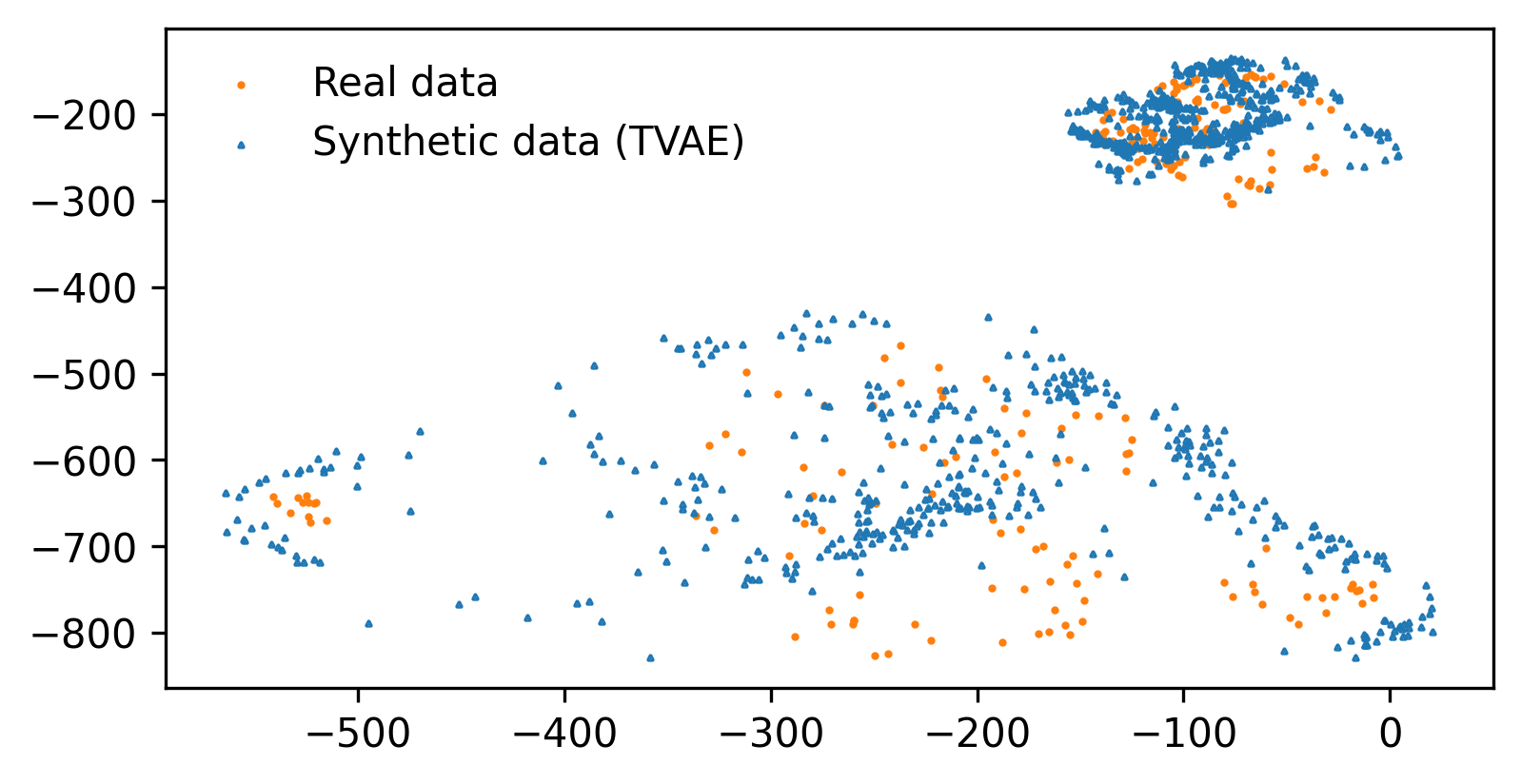}}
	\subfigure[Copula GAN]{\includegraphics[width=.49\textwidth]{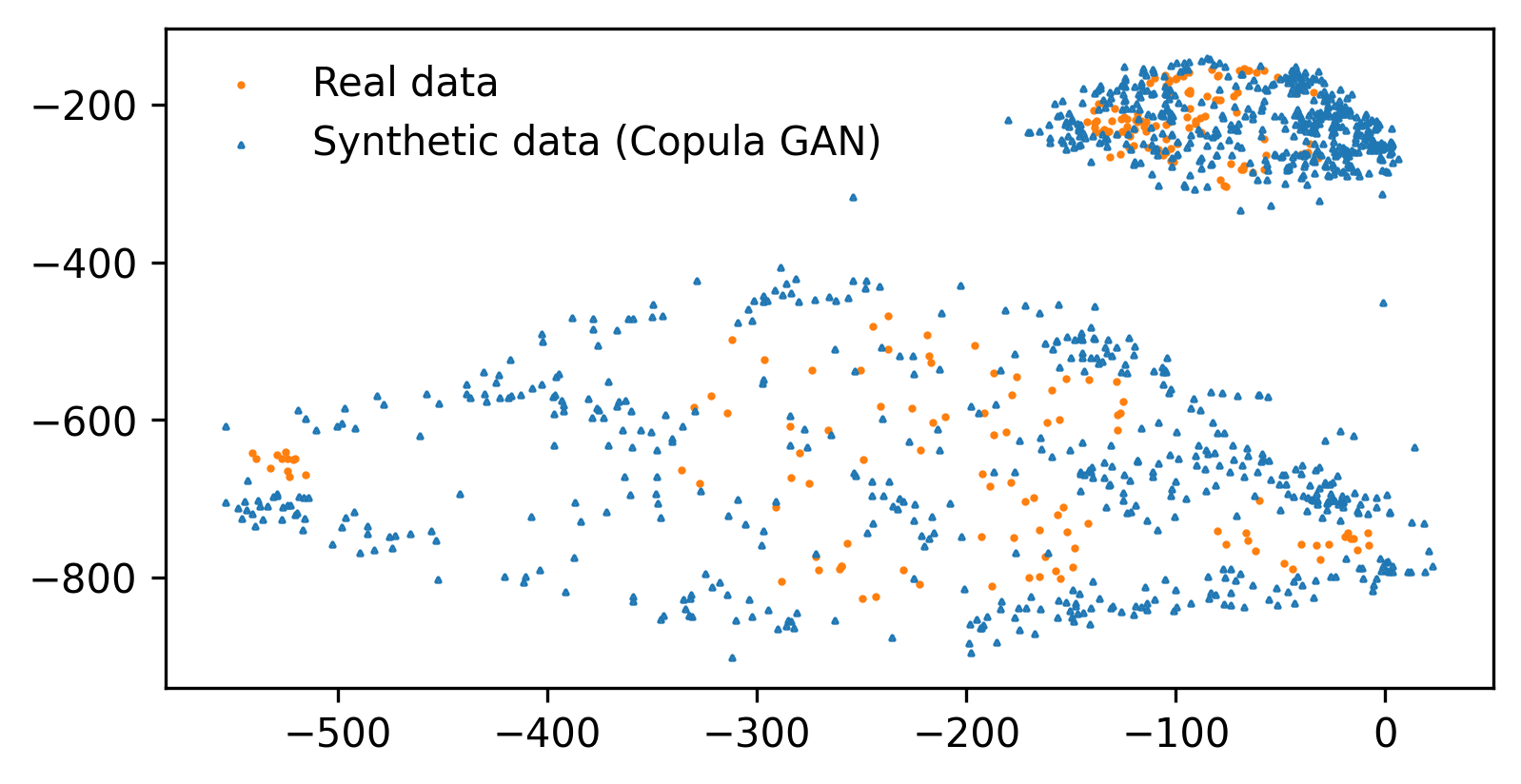}}
	\caption{2D visualization of generated synthetic data for obesity risk dataset}\label{Fig:Obesity}
\end{figure}

Figure \ref{Fig:Obesity} presents a 2D visualization of real dataset and each generated synthetic dataset. CTGAN and TVAE produce more representative synthetic dataset of the real dataset since the generated instances are uniformly distributed, simulating that better mimic the global characteristics of the real data.  In contrast, Gaussian Copula and CopulaGAN present bias and tend not to give sufficient focus or omitting some regions of the data space, which may lead to less reliable synthetic data.

\begin{table}[!ht]
	\setlength{\tabcolsep}{5pt}
	\renewcommand{\arraystretch}{1.}
	\centering %
	\begin{tabular}{l|cccc}
		\toprule
		Evaluation test        & GaussianCopula &   CTGAN    &   TVAE   & CopulaGAN \\ 
		\midrule
		Wasserstein-Cramer's V &     37.41      & \bf{29.85} &  35.32   &   35.41   \\
		Novelty                &    \bf{0.0}    &  \bf{0.0}  & \bf{0.0} & \bf{0.0}  \\
		Domain classifier      &     0.784      & \bf{0.764} &  0.770   &   0.773   \\ 
		Anomaly detection      &     0.005      &   0.011    & \bf{0.0} &   0.021   \\
		\bottomrule
	\end{tabular}
	\caption{Score on evaluation tests of Gaussian Copula, GMM, CTGAN, TVAE and CopulaGAN models for obesity risk dataset}\label{Table:Obesity results}
	\setlength{\tabcolsep}{10pt}
	\renewcommand{\arraystretch}{1.}
	\centering %
	\begin{tabular}{l|ccc}
		\toprule
		Model    & Friedman & \multicolumn{2}{c}{\underline{Finner post-hoc test}}\\
		& Ranking  & $APV_i$ & $H_0$\\
		\midrule
		CTGAN            & 5.625   &    $-$     &     $-$      \\
		TVAE             & 7.125   & 0.133614   &    Failed to reject \\
		CopulaGAN        & 9.825   & 0.000030   &    Rejected  \\
		Gaussian Copula  & 11.37   & 0.000000   &    Rejected  \\
		\bottomrule 
	\end{tabular}
	\caption{Statistical analysis provided by the proposed evaluation framework for obesity risk dataset}\label{Table:Obesity}
\end{table}

\section{Conclusions and future research}\label{Sec:6}

The evaluation of the quality of synthetic data constitutes a vital role in many real-world applications offering a cost-efficient approach for enhancing data augmentation while addressing concerns related to sensitive data privacy as well as improving machine learning models performance. 

In this work, we introduced a new statistically-based framework for comparing the performance of synthetic data generation models. The proposed framework employs several multivariate evaluation tests for measuring the quality of generated data of each model and then, a statistical analysis is conduced for ranking the evaluated models and examine for significant differences in their performance using FAR and Finner post-hoc tests. A considerable advantage of the proposed approach is that it is able to provide strong theoretical and statistical results about the models' ranking and the overall evaluation process. In addition, the proposed framework offers  flexibility and adaptivity in sense that new tests can be easily integrated and it can be employed for evaluating the quality of any generated synthetic dataset, respectively. The use case scenarios on two real-world datasets demonstrated the applicability of the proposed framework and its ability for evaluating state-of-the-art synthetic data generation models. It is also worth mentioning that we examined the case where the real data is unlabeled, which considerable increases the complexity of the evaluation process. 

Clearly, in case a evaluation test consists of primary importance, then the proposed evaluation framework can act a supplementary role to decision making process. In other case, the proposed framework is able to conduct a direct comparison on the models' ability to generate high quality synthetic data, especially when the evaluation tests' conclusions are conflict. Its primary aim is to provide a statistical-based approach for addressing the problem of evaluating the quality of synthetic datasets, especially in cases where the interpretation of evaluation tests results require advanced knowledge and/or the utilized evaluation tests generated conflicting conclusions; therefore, addresses the limitations of existing evaluation methodologies.

A limitation of this work is that the use scenarios concerned tabular data. In our future work, we intent to apply the proposed approach on image synthesis applications \cite{mao2021generative}. 
This expansion focuses on further exploring the adaptability and effectiveness of the proposed framework across diverse data modalities as well as gaining insights related to its  applicability in scenarios where the data exhibits different structures and characteristics.

Finally, it is worth mentioning that the selected tests in the proposed framework do not constitute a conclusive list. An extension may introduce new attributes and other criteria, which were not included in our research and may potentially influence the final conclusions/findings. Along this line, our future work is concentrated on including other statistical tests such as Kolmogorov-Smirov test \cite{justel1997multivariate}  and Kullback–Leibler divergence test \cite{kullback1951information}. Furthermore, an interesting idea is to update the proposed framework by introducing different weights to each component test. In this way, we will be able to assign increased importance to some tests during the statistical analysis, rather than considering their impact as equivalent.

\noindent \textbf{Acknowledgements} This work has been conducted under the auspices of the General Secretariat for Research Innovation, for the certification of research and development expenditure programme, under No. 82216, project Synthetic data synthesis - Datafication of social data.


\end{document}